\documentclass[10pt,twocolumn,letterpaper]{article}

\usepackage{ijcb}
\usepackage{times}
\usepackage{epsfig}
\usepackage{caption,subcaption,graphicx}
\usepackage{amsmath}
\usepackage{amssymb}

\usepackage{textcomp}
\usepackage{xcolor}
\usepackage{mathrsfs}
\usepackage{booktabs} 

\makeatletter
\@namedef{ver@everyshi.sty}{}
\makeatother
\usepackage{dblfloatfix}
\usepackage{bigstrut}
\usepackage{multirow}
\usepackage{color}
\usepackage{caption}
\usepackage{fancyhdr}
\usepackage{listings} 
\usepackage{changepage}
\usepackage{pgfplots}
\usepackage{balance}
\usepackage{enumitem}

\usepackage{breqn}
\usepackage{dblfloatfix}

\ijcbfinalcopy 

\makeatletter
\def\ps@IEEEtitlepagestyle{
\def\@oddfoot{\mycopyrightnotice}
\def\@evenfoot{}
}
\def\mycopyrightnotice{
{\hfill \footnotesize 978-1-7281-9186-7/20/\$31.00 \copyright 2020 IEEE\hfill}
}
\makeatother

\ifijcbfinal\fi
\begin{document}

\title{\textbf{On the Influence of Ageing on Face Morph Attacks: Vulnerability and Detection}}

\author{Sushma Venkatesh \quad  Kiran Raja  \quad Raghavendra Ramachandra  \quad Christoph Busch\\
Norwegian University of Science and Technology (NTNU), Norway\\ 
E-mail:	\{\tt\small sushma.venkatesh;kiran.raja;raghavendra.ramachandra;christoph.busch\} @ntnu.no\\
}

\maketitle
\thispagestyle{empty}

\begin{abstract}
Face morphing attacks have raised critical concerns as they demonstrate a new vulnerability of Face Recognition Systems (FRS), which are widely deployed in border control applications.  The face morphing process uses the images from multiple data subjects and performs an image blending operation to generate a morphed image of high quality. The generated morphed image exhibits similar visual characteristics corresponding to the biometric characteristics of the data subjects that contributed to the composite image and thus making it difficult for both humans and FRS, to detect such attacks. In this paper, we report a systematic investigation on the vulnerability of the Commercial-Off-The-Shelf (COTS) FRS when morphed images under the influence of ageing are presented. To this extent, we have introduced a new morphed face dataset with ageing derived from the publicly available MORPH II face dataset, which we refer to as MorphAge dataset. The dataset has two bins based on age intervals, the first bin - MorphAge-I dataset has 1002 unique data subjects with the age variation of 1 year to 2 years while the MorphAge-II dataset consists of 516 data subjects whose age intervals are from 2 years to 5 years. To effectively evaluate the vulnerability for morphing attacks, we also introduce a new evaluation metric, namely the Fully Mated Morphed Presentation Match Rate (FMMPMR), to quantify the vulnerability effectively in a realistic scenario. Extensive experiments are carried out  using two different COTS FRS (COTS I  Cognitec FaceVACS-SDK Version 9.4.2
and COTS II - Neurotechnology version 10.0) to quantify the vulnerability with ageing. Further, we also evaluate five different Morph Attack Detection (MAD) techniques to benchmark their detection performance with respect to ageing. 
\end{abstract}

\let\thefootnote\relax\footnotetext{\mycopyrightnotice}
\section{Introduction}
\label{sec:introduction}

\begin{figure}[htp]
	\centering
	\includegraphics[width=1\linewidth]{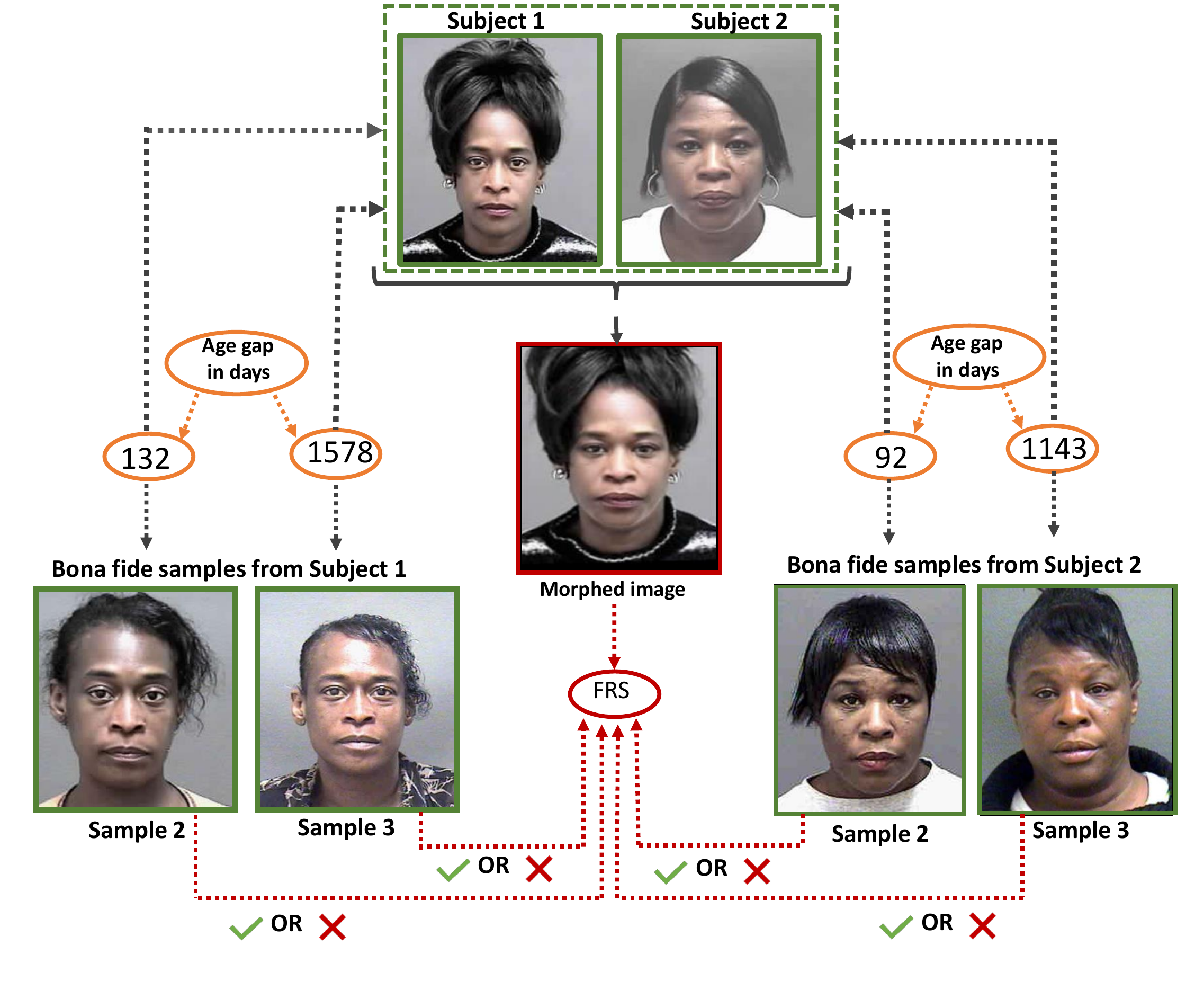}
	\vspace{-3mm}
	\caption{Illustration of the influence of ageing on face morphing}
	\label{fig:Intro}
\end{figure}

Facial characteristics have been well explored for identifying and verifying individuals and numerous biometric systems have been deployed in operational applications for many years \cite{jain2007handbook, jain2011handbook}. The preference towards face based biometric systems is founded on multiple factors such as ease of capture of facial characteristic without invasive imaging, capturing at a stand-off distance both in semi-cooperative (voluntary identification/verification) and uncooperative scenarios (surveillance) \cite{klare2015pushing, cao2018vggface2, kemelmacher2016megaface}. While many of the breakthrough articles detailing iris and vein recognition systems have shown impeccable accuracy with very low false accepts and false rejects, those systems suffer from highly constrained image capturing processes. In order to reach the performance of such iris and vein recognition systems, face biometrics has seen benefits from recent algorithmic advancements, which was focused on features that have been engineered in a robust manner \cite{lu2015learning, lu2015simultaneous, chen2017robust}, and pre-processing that has been improvised \cite{punnappurath2015face} by including end-to-end learning using Deep Neural Networks (DNN) even in large scale applications \cite{kemelmacher2016megaface, ranjan2017hyperface}.

Such attractive and inherent advantages of the face modality have led to a wider deployment of Face Recognition Systems (FRS) in passport issuance processes, visa management, identity management and Automated Border Control (ABC). Despite the high accuracy and convenience of face biometrics, FRS systems are impeded by various factors such as ageing \cite{park2010age,jain2011face}, partial face availability \cite{jain2011face} and also imperilled by various attacks that include presentation attacks (spoofing-attacks) with print, display or silicon mask attack instruments \cite{marcel2014handbook}, make-up attacks \cite{wang2016recognizing, bharati2016detecting}, coverted mask attacks \cite{erdogmus2013spoofing}, morphing attacks \cite{ferrara2014magic}, database level attacks \cite{jain2005biometric, biggio2013poisoning} and comparison level attacks \cite{galbally2010vulnerability, jain201650}. While many of the attacks have been addressed through mitigation measures over the period of time, we focus on recently surfaced face morphing attacks \cite{ferrara2014magic, Raghavendra2016} in this work. Despite some of the recent works proposing measures to mitigate these attacks through various approaches \cite{scherhag2019face, ferrara2018face, ferrara2017face, Raghavendra2016} a number of covariates are reported to impact the attack detection performance. A list of covariates impacting the performance of morphing attack detection include the techniques used to generate the morphed image \cite{ferrara2014magic}, the configuration of the print-scan pipeline \cite{raghavendra2017transferable}, factors of age and ethnicity \cite{scherhag2017biometric} among many other unknown factors. With a clear introspection of the existing works, we observe that both the FRS vulnerability and also the Morphing Attack Detection (MAD) performance under variation of age is not studied in the context of morphing attacks, despite the fact that the issue was pointed out already in the early works in this domain \cite{scherhag2017biometric, ferrara2017face}.

Starting with this observation, we focus in this work on establishing the impact of ageing on morphing attacks by carefully studying the vulnerability of FRS and MAD performance of currently reported MAD algorithms under the influence of ageing. The key motivation stems from earlier works who have disentangled the impact of ageing on face recognition systems with respect to recognition performance \cite{park2010age,jain2011face, best2017longitudinal} and a number of works that have proposed approaches to handle the associated performance limitations \cite{park2010age, li2011discriminative, gong2013hidden, li2018distance, best2017longitudinal}. We therefore provide a brief overview of impact of ageing in the subsequent section and thereafter illustrate the impact of ageing specifically for morphing. Further, we focus our work on investigating the impact using the digital images alone due to two primary factors: (i) many countries across the world allow to upload digital images via web-portal for passport renewal and visa issuance, and (ii) to align our works with recent studies focusing on digital MAD \cite{ferrara2017face}.

\subsection{Facial Ageing}
Facial ageing is a commonly observed phenotype of human ageing, which is visibly seen. Despite the complexity of understanding the characteristic changes associated with the facial ageing, a number of works have reported the role of skin and soft tissues and their impact on visible changes of facial appearance \cite{farkas2013science}.  Complementary works have demonstrated the role of loss of facial bone volume to contribute to facial appearance under ageing progress \cite{shaw2011aging}. As it can be deduced, facial ageing being a complex process involving soft tissues and skeletal structure changes, it is influenced by many factors, such as exposure to sunlight and body weight among others. As an additional factor, large variations in facial ageing across individuals and ethnic populations can further be observed \cite{panis2016overview}. While in face recognition, the main differences in exterior facial structure making individuals distinguishable from each other allows recognition analysis to achieve high identification accuracy, a longitudinal study of the same face over a period of time has shown to challenge the accuracy \cite{best2017longitudinal}.

\subsection{Facial Ageing and Morphing Attacks}
\label{sec:FAMA}
Under the observation of complex changes of facial appearance, which bring down the recognition accuracy of FRS unless proper measures are taken, our assertion is that the effect and impact on morphing attacks may change. For electronic Machine Readable Travel Documents (MRTDs) a typical life-cycle of 10 years is recommend \cite{erickson2013passport} meaning that the drastic changes in facial appearance must be tolerated as intra-class variance during that life-cycle, while up to now the impact of morphing and its correlation with the progressing of the potentially morphed reference image in this life-cycle, has neither been considered nor investigated. Initial studies on morphing attacks have demonstrated the ability to fool a human expert (i.e. trained border guards) with morphed facial images. The changes of facial appearance, which are caused by ageing, are illustrated in Figure \ref{fig:Intro}. Our assertion is to validate the impact of ageing and thus we formulate three specific research questions:
\begin{itemize}[leftmargin=*,noitemsep, topsep=0pt,parsep=0pt,partopsep=0pt]
    \item How vulnerable are COTS FRS when a composite morph image is enrolled and is after a period of ageing probed against a live image from one of the contributing subjects?  
    \item Do current Morphing Attack Detection (MAD) algorithms scale-up to detect such attacks under the influence of ageing? 
    \item What is the impact of different alpha (or blending, morphing) factors used to generate the morphed image under the constraint of ageing, specifically with respect to MAD? 
\end{itemize}
We address each of these questions in a systematic manner through our contributions. We focus in this work to first establish the impact on FRS through an extensive empirical evaluation. While a detailed study of appearance change is more of a cognitive study, it is beyond the scope of the current work.

\subsection{Contributions of Our Work}
While the hypothesis is well justified, we also note that there exists no database with morphing and ageing according to the current literature. With such a caveat, we focus on first creating a database to facilitate and validate our assertion. 
\begin{itemize}[leftmargin=*,noitemsep, topsep=0pt,parsep=0pt,partopsep=0pt]
    \item The first key contribution is the creation of a (moderately) large-scale database of morphed faces with ageing covariate by employing the MORPH II non-commercial face dataset \cite{MORPHII_DB}, which is hereafter referred  as MorphAge Database.
    \item We investigate the vulnerability of FRS to such attacks by employing two widely used Commercial-Off-The-Shelf (COTS) FRS systems. This contribution not only helps in verifying our assertion but also validates the usefulness of the newly created database. Further, we also investigate the role of alpha (or blending, morphing) factor  (with $\alpha =$ 0.3. 0.5 and 0.7) while analysing the vulnerability under ageing.  
    \item As a third contribution, we employ a set of recently reported morphing attack detection algorithms to benchmark detection performance and thereby identify the impediments if any.
\end{itemize}

In the rest of the paper, we first provide details on the newly constructed database in Section~\ref{sec:database} and in Section~\ref{sec:vulnerability} we investigate the vulnerability of FRS using two COTS FRS. Further, the benchmarking of morphing attack detection systems is detailed in Section~\ref{sec:performance} while the key observations and conclusions are reported in Section~\ref{sec:conclusion}.

\section{MorphAge Database Construction}
\label{sec:database}
To effectively study the influence of ageing on face morphing vulnerability and morph detection, we introduce a new dataset, which is derived from the MORPH II non-commercial dataset \cite{MORPHII_DB} that is publicly available. The MORPH II dataset consists of a total of $55000$ unique samples captured from $13000$ data subjects. The images are captured over the time span from 2003 to 2007. The age of the subjects varies from 16 to 77 years. The dataset consists of male and female subjects with different ethnicity (African, European, Asian, Hispanic). In this work, we choose the MORPH II dataset motivated by the large number of subjects, the quality of the captured data and the variation in age for one and the same subject across different capture sessions. 

The newly constructed MorphAge dataset is binned in two age groups from MORPH II dataset. The first bin - Age Group (MorphAge-I) consists of 1002 unique data subjects with a gender distribution of 143 female and 859 male subjects. For each data subject, three different samples are chosen such that the first session corresponds to the high quality data capture (younger age), second session corresponds to the aged capture of 1-8 months from first session and third session corresponds to the aged capture of same subject between 1-2 years from first session. The second bin - Age Group (MorphAge-II) is comprised of 516 unique data subjects sub-sampled from the MORPH II dataset with 62 female and 454 male data subjects. Each data subject was captured in three different sessions. The first session corresponds to the high quality data capture (younger age), the second session corresponds again to a time lapse of 1-8 months from the first session and the third session corresponds to an aged capture of 2 years up to 5 years after the first session.

\begin{figure}[htp]
	\centering
	\includegraphics[width=1\linewidth]{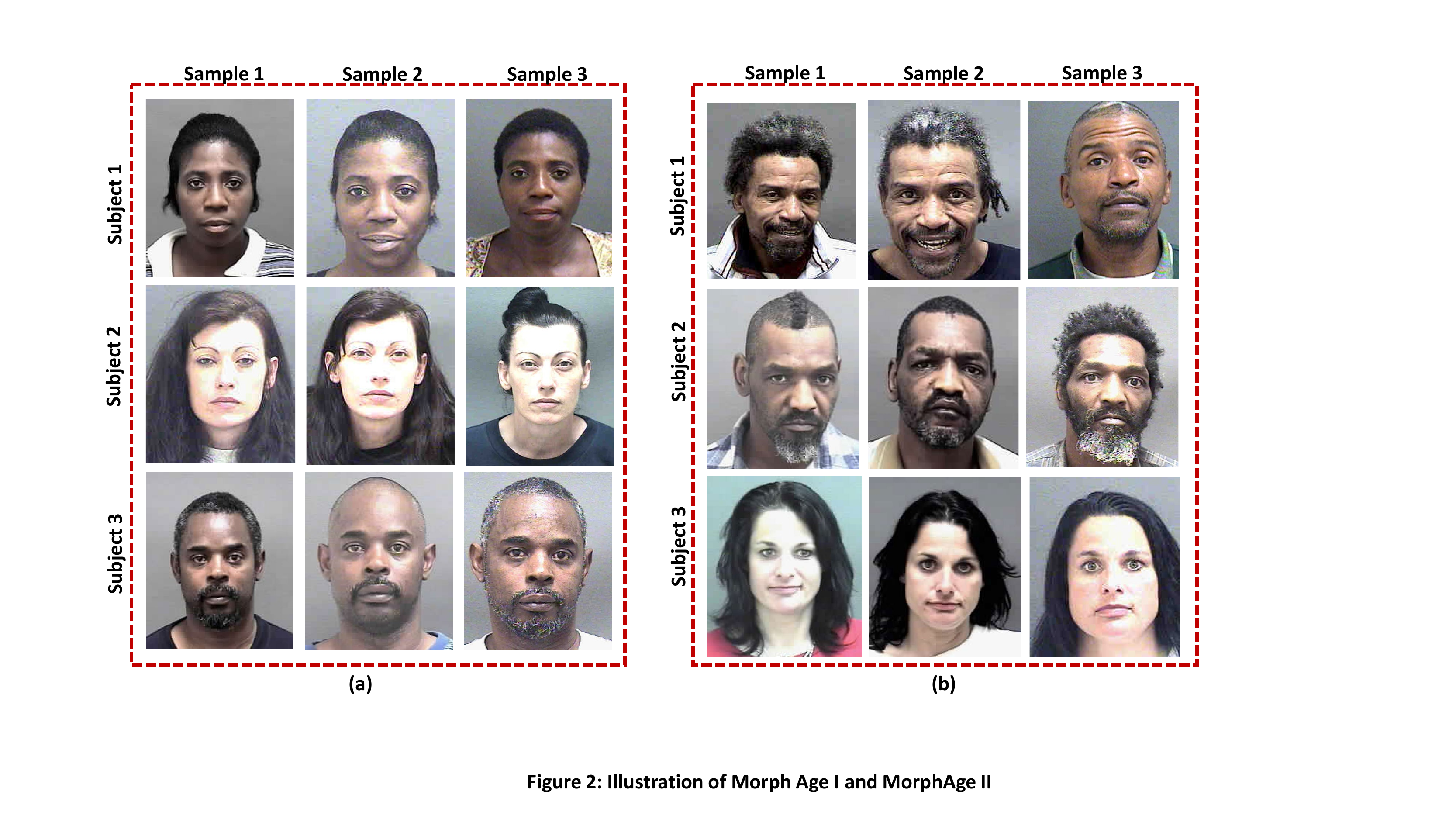}
	\vspace{-3mm}
	\caption{Illustration of sample images from newly constructed MorphAge dataset (a) MorphAge-I (1 year to 2 years) (b) MorphAge-II (2 years to 5 years)}
	\label{fig:DBIms1}
\end{figure}
\begin{table}[htbp]
	\centering
	\caption{Statistics of bona fide and morphed images in MorphAge Database}
	\resizebox{1\linewidth}{!}{
		\begin{tabular}{lcccc}
			\hline
			\textbf{Session} & \textbf{Dev} & \textbf{Training} & \textbf{Testing} & \textbf{Total} \bigstrut\\
			\hline
			\hline
			\multicolumn{5}{c}{MorphAge-I Subset} \bigstrut\\
			\hline
			\hline
			Session 1  & 251   & 500   & 251   & 1002 \bigstrut\\
			(used for morphing)& & & & \\
			\hline
			Session 2 & 251   & 500   & 251   & 1002\bigstrut\\
			(used for vulnerability) & & & & \\
			\hline
			Session 3  & 251   & 500   & 251   & 1002 \bigstrut\\
			(with age difference)& & & & \\
			\hline
			Morphed Images & 1980  & 6614  & 1944  & 10538 \bigstrut\\
			\hline
			\hline
			\multicolumn{5}{c}{MorphAge-II Subset} \bigstrut\\
			\hline
			\hline
			Session 1  & 130   & 257   & 129   & 516 \\
			(used for morphing) & & & & \\
			\hline
			Session 2 & 130   & 257   & 129   & 516 \\
			(used for vulnerability) & & & & \\
			\hline
			Session 3 & 130   & 257   & 129   & 516 \\
			(with age difference)& & & & \\
			\hline
			Morphed Images & 648   & 2310  & 809   & 3767 \bigstrut\\
			(with different morphing factors) & & & & \\
			\hline
		\end{tabular}%
	}
	\vspace{-4mm}
	\label{tab:DB_stats1}%
\end{table}
\subsection{MorphAge-I and MorphAge-II - Bonafide Set}
In both bins, i.e. MorphAge-I and MorphAge-II, we select for each data subject three samples (one of each session) such that the \textsl{first session sample is used only to generate the morphing image, the sample from the second session is used as bona fide sample in the morph attack detection experiments and the third session to analyze the vulnerability of the commercial FRS}. 
As seen from the Figure \ref{fig:DBIms1}, the facial appearance changes significantly with the increasing age which cannot be modelled geometrically or morphologically for any particular ethnicity or age group for both the bins (MorphAge-I and MorphAge-II).

\begin{figure}[htp]
	\centering
	\includegraphics[width=1\linewidth]{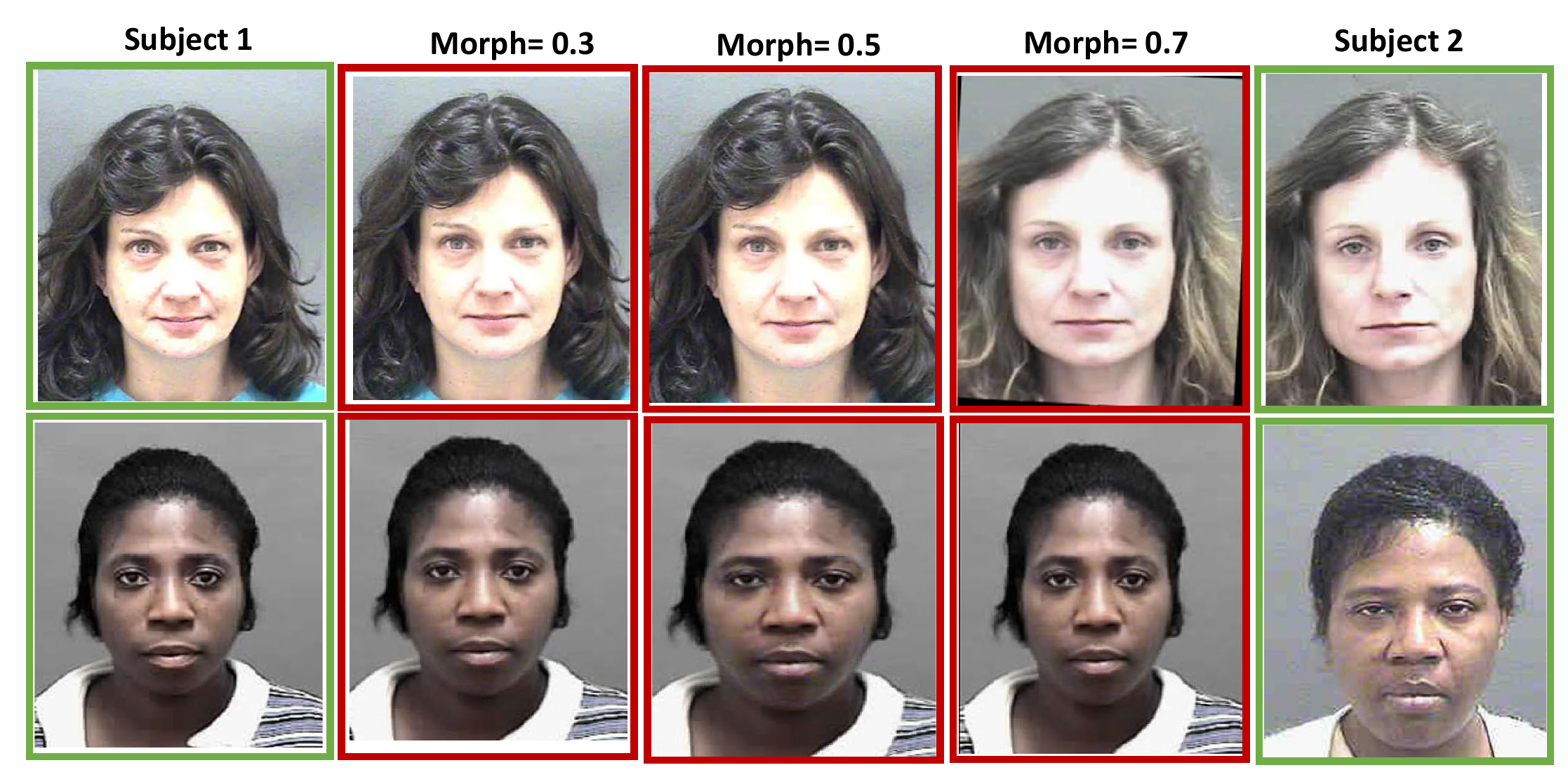}
	\vspace{-3mm}
	\caption{Example of generated morphed images}
	\label{fig:DBIms2}
\end{figure}
\subsection{MorphAge-I and MorphAge-II - Morphed Image Set}
To generate the morphed image datasets for the subjects represented in our newly constructed dataset, we have used the face morph generation tool from Ferrara et al. \cite{UBO_Morphing_Tool} \cite{Ferrara2016}, which is based on facial landmarks based warping and weighted linear blending to generate a high quality morphed image. We particularly, choose this technique for morphing generation over other type of generators based on GAN \cite{MorGAN} by considering:  (1) high quality of the generated morphed images, in order to establish a significant threat to the tested commercial FRS \cite{UBO_Morphing_Tool} (2) high quality of generated morphed image, such that the submitted images are considered compliant with the requirements in the ICAO standards and (3) feasibility to create the morphed images with various blending and warping factors.

\begin{figure*}[!tbp]
  \centering
  \begin{minipage}[b]{0.7\textwidth}
	\includegraphics[width=1\linewidth]{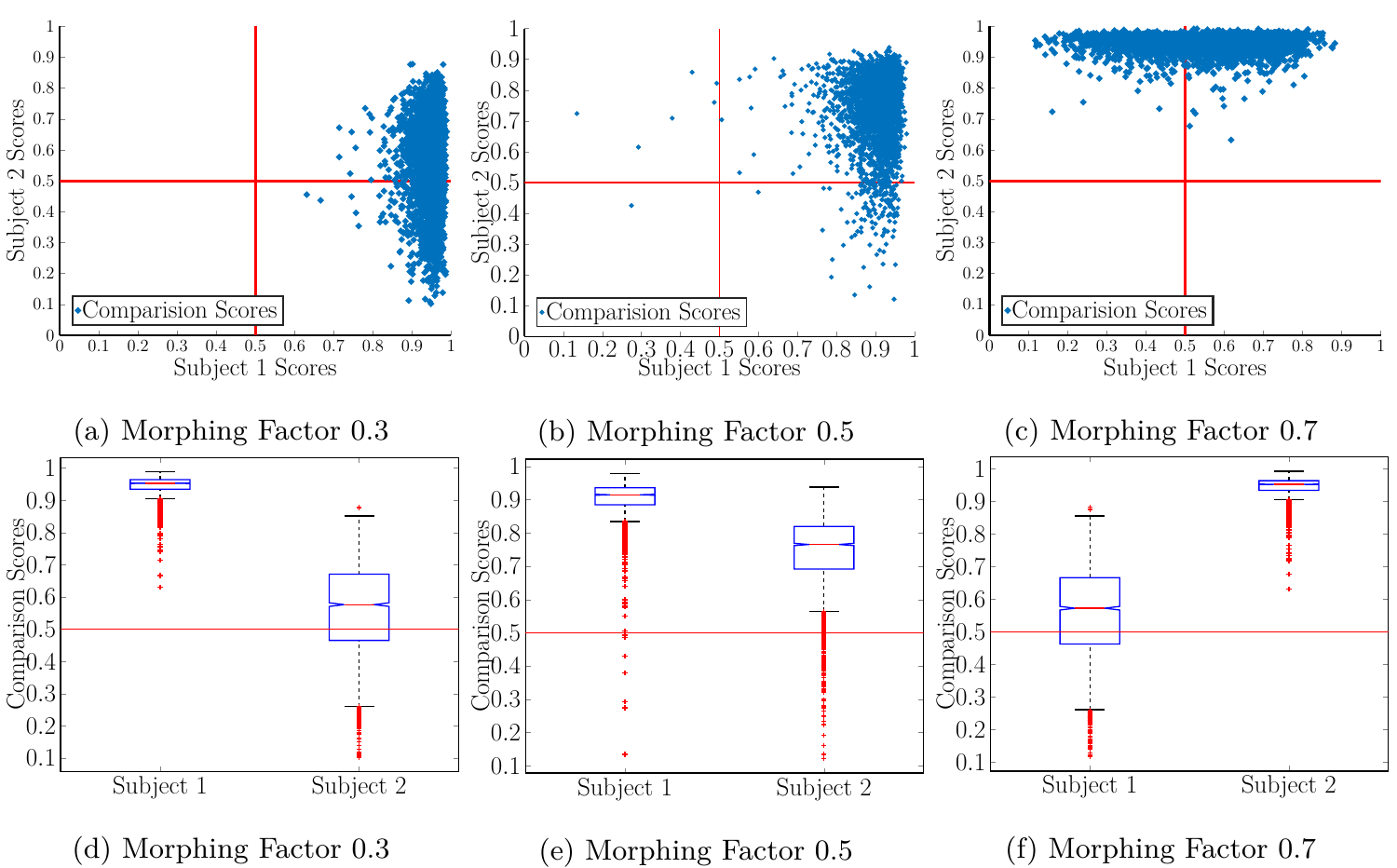}
	\vspace{-3mm}
	\caption{Scatter and Box plots obtained using COTS-I FRS on MorphAge-I dataset}
	\label{fig:CogAGI}
  \end{minipage}
  \hfill
  \begin{minipage}[b]{0.7\textwidth}
		\includegraphics[width=1\linewidth]{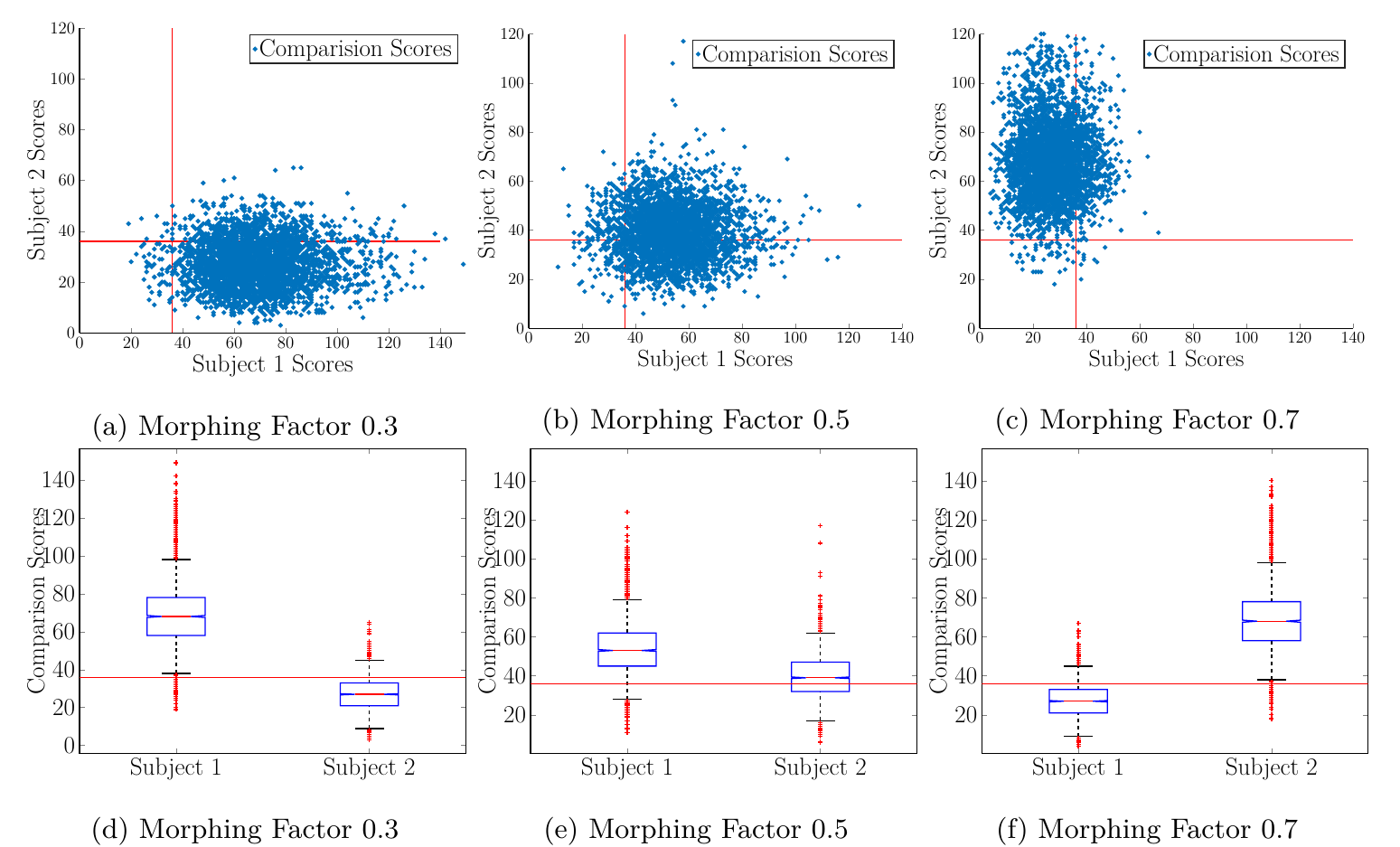}
	\vspace{-3mm}
	\caption{Scatter and Box plots obtained using COTS-II FRS on MorphAge-I dataset}
	\label{fig:NeuAGI}
  \end{minipage}
\end{figure*}

In this work, the morphing process is carried out between only two data subjects by considering its use-case in a real-life scenario where typically one criminal morphs his/her face image with the image of an accomplice. To carefully select the pair of images for the morphing process, we use the 
COTS-I FRS, which is widely used in Automated Border Control installations. Through the FRS, a set of similarity scores is obtained between the probe image of a selected data subject against the reference images of all data subjects. We then choose the pair of images that are successfully verified at FMR = $0.1\%$ with high scores to retain a high degree of similarity between two constituting subjects for the morphed image. Additional care is exercised not to combine data subjects with different genders and also to separate the data subject into three independent groups such as non-overlapping training, testing and development sets \cite{scherhag2017biometric,UScherhag_Survey}. For a selected image pair, we generate three morphed images at three different morphing (or blending) factors $\alpha = 0.3, 0.5, 0.7$ to obtain insights with regard to the impact of ageing at different blending factors. Figure \ref{fig:DBIms2} shows the example of morphed face images with three different blending factor within our MorphAge dataset. 

Table \ref{tab:DB_stats1} presents the statistics of the generated dataset corresponding to the two bins - MorphAge-I and MorphAge-II. Further, in order to evaluate the MAD performance, we have divided the whole datasets into three independent and non-overlapping subsets for training, development and testing. The training subset is used purely to train the MAD techniques, the development subset is used to optimize and adjust the operating threshold for the MAD techniques and finally the testing subset is solely used to analyze the detection performance obtained at the optimal threshold. 

\section{Vulnerability Analysis}
\label{sec:vulnerability}
In this section, we present the vulnerability analysis of the FRS, when confronted with the morphed images under variation of age. To this extent, we employ two different COTS Face Recognition Systems (FRS) namely, COTS-I Cognitec\footnote{Outcome not necessarily constitutes the best the algorithm can do.} FaceVACS-SDK Version 9.4.2 and COTS-II Neurotechnology Version 10.0. 
To effectively measure the vulnerability of the FRS against morphed face samples, we set a realistic constraint that all contributing data subjects (in our case two) must exceed the verification threshold of the FRS. Further, in this work, we set the operating threshold of both COTS FRS to FAR = 0.1$\%$ following the guidelines of FRONTEX \cite{EU-Frontex-BestPracticeABC-2015} for automated border control. Thus, we coin the new realistic constraint using a new vulnerability metric as \textsl{Fully Mated Morphed Presentation Match Rate (FMMPMR)} that can be computed as: 
%
%

\begin{dmath}
	FMMPMR = \frac{1}{P} \sum_{M,P}^{} {(S1_{M}^{P} > \tau) \&\& (S2_{M}^{P} > \tau) \\ \ldots \&\&  (Sk_{M}^{P} > \tau)}
	\label{Eqa:FMMPMR}
\end{dmath}
Where $P = {1, 2, \ldots, p}$ represent the number of attempts made by presenting all the probe images from the contributing subject against $M^{th}$ morphed image,  $K = {1, 2, \ldots, k}$ represents the number of contributing data subjects to the  constitution of the generated morphed image (in our case  $K=2$), $Sk_{M}^{P}$ represents the comparison score of the $K^{th}$ contributing subject obtained with $P^{th}$ attempt (in our case the $P^{th}$ probe image from the dataset) corresponding to $M^{th}$ morph image and $\tau$ represents the threshold value corresponding to FAR = 0.1$\%$. 

We have employed the new metric FMMPMR considering the fact that the existing vulnerability metric MMPMR\cite{scherhag2017biometric} accounts only for the morphed images getting verified with the contributing subjects without taking into account the number of attempts. However, the new metric FMMPMR overcomes this drawback and considers each and every attempt a morphed image gets verified with the pair of contributing subjects, i.e., reflecting the actual vulnerability of a FRS.

\begin{table}[htbp]
  \centering
  \caption{Vulnerability analysis: FMMPMR (\%)}
  \resizebox{0.94\linewidth}{!}{
    \begin{tabular}{|c|c|c|c|c|}
    \hline
    \multirow{3}[6]{*}{Morphing factor ($\alpha$)} & \multicolumn{4}{c|}{\textbf{FMMPMR}(\%)} \bigstrut\\
\cline{2-5}    \multicolumn{1}{|c|}{} & \multicolumn{2}{c|}{MorphAge-I} & \multicolumn{2}{c|}{MorphAge-II} \bigstrut\\
\cline{2-5}    \multicolumn{1}{|c|}{} & COTS-I & COTS-II & \multicolumn{1}{c|}{COTS-I} & COTS-II \bigstrut\\
    \hline
    0.3   & 66.24 & 18.42  & \multicolumn{1}{c|}{58.47} & 17.29 \bigstrut\\
    \hline
    0.5   & 95.07 & 56.96 & \multicolumn{1}{c|}{93.81} & 51.27 \bigstrut\\
    \hline
    0.7   & 67.32 & 18.21 & 58.18   & 15.61 \bigstrut\\
    \hline
    \end{tabular}%
    }
  \label{tab:FMMPMR}%
\end{table}%

\begin{table}[htbp]
	\centering
	\caption{Experiment-I: Quantitative performance of the MAD techniques on MorphAge-I}
	\resizebox{1\linewidth}{!}{
		\begin{tabular}{|c|c|c|c|c|c|}
			\hline
			\multirow{3}[6]{*}{\textbf{Algorithm}} & \textbf{Development} & \multicolumn{4}{c|}{\textbf{Testing set}} \bigstrut\\
			 &  \textbf{Set} &  \multicolumn{4}{c|}{} \bigstrut\\
            \cline{2-6} & \multirow{2}{*}{EER (\%)} & \multirow{2}{*}{EER (\%)} & \multicolumn{3}{c|}{BPCER (\%) @ APCER (\%) =} \bigstrut\\
            \cline{4-6}  &   &   & 1 & 5 & 10 \bigstrut\\
			\hline
			\multicolumn{6}{|c|}{Morphing factor ($\alpha$) 0.3} \bigstrut\\
			\hline
			LBP-SVM \cite{Raghavendra2016, Luuk_FMdetect2018, DBLP:journals/corr/abs-1901-08811, Scher2017}  & 28.14 & 35.11 & 84.4 & 68.8 & 56.8 \bigstrut\\
			\hline
			BSIF-SVM \cite{Raghavendra2016, Scher2017} & 31.82 & 37.59 & 98.8 & 90 & 73.2 \bigstrut\\
			\hline
			HOG-SVM cite{Scher2017} & 32.09 & 33.51 & 84.4 & 63.6 & 53.6 \bigstrut\\
			\hline
			AlexNet-SVM \cite{DBLP:journals/corr/abs-1901-08811, RagCVIP2018, DeepLearning_JanaDittman} & 4.38 & 2 & 7.2 & 3.2 & 0.8 \bigstrut\\
			\hline
			Color Denoising \cite{Sushma_IPTA2019}& 1.63 & 3.65 & 5.2 & 0.4 & 0.4 \bigstrut\\
			\hline
			\multicolumn{6}{|c|}{Morphing factor ($\alpha$) 0.5} \bigstrut\\
			\hline
			LBP-SVM \cite{Raghavendra2016, Luuk_FMdetect2018, DBLP:journals/corr/abs-1901-08811, Scher2017}  & 27.82 & 33.76 & 75.2 & 59.2 & 57.2 \bigstrut\\
			\hline
			BSIF-SVM \cite{Raghavendra2016, Scher2017} & 31.82 & 36.9 & 98.8 & 89.21 & 73.6 \bigstrut\\
			\hline
			HOG-SVM \cite{Scher2017} & 30.73 & 34.1 & 81.2 & 63.2 & 56.8 \bigstrut\\
			\hline
			AlexNet-SVM \cite{DBLP:journals/corr/abs-1901-08811, RagCVIP2018, DeepLearning_JanaDittman} &  3.18 & 2.01  & 4.12  &0    & 0 \bigstrut\\
			\hline
			Color Denoising \cite{Sushma_IPTA2019}& 1.63 & 1.21 & 7.6 & 0.4 & 0 \bigstrut\\
			\hline
			\multicolumn{6}{|c|}{Morphing factor ($\alpha$) 0.7} \bigstrut\\
			\hline
			LBP-SVM \cite{Raghavendra2016, Luuk_FMdetect2018, DBLP:journals/corr/abs-1901-08811, Scher2017} & 28.86 & 34.92 & 88.4 & 66.8 & 57.2 \bigstrut\\
			\hline
			BSIF-SVM \cite{Raghavendra2016, Scher2017} & 31.9 & 37.98 & 98.8 & 88 & 73.2 \bigstrut\\
			\hline
			HOG-SVM \cite{Scher2017}  & 32.98 & 33.38 & 80 & 62.8 & 57.2 \bigstrut\\
			\hline
			AlexNet-SVM \cite{DBLP:journals/corr/abs-1901-08811, RagCVIP2018, DeepLearning_JanaDittman} & 5.08 & 2.78 & 5.6 & 2 & 0 \bigstrut\\
			\hline
			Color Denoising \cite{Sushma_IPTA2019} & 2.75 & 2.43 & 13.2 & 2 & 0.4 \bigstrut\\
			\hline
		\end{tabular}%
	}
	\label{tab:AGI_EXP-III}%
\end{table}%



\begin{figure*}[!tbp]
  \centering
  \begin{minipage}[b]{0.7\textwidth}
	\includegraphics[width=1\linewidth]{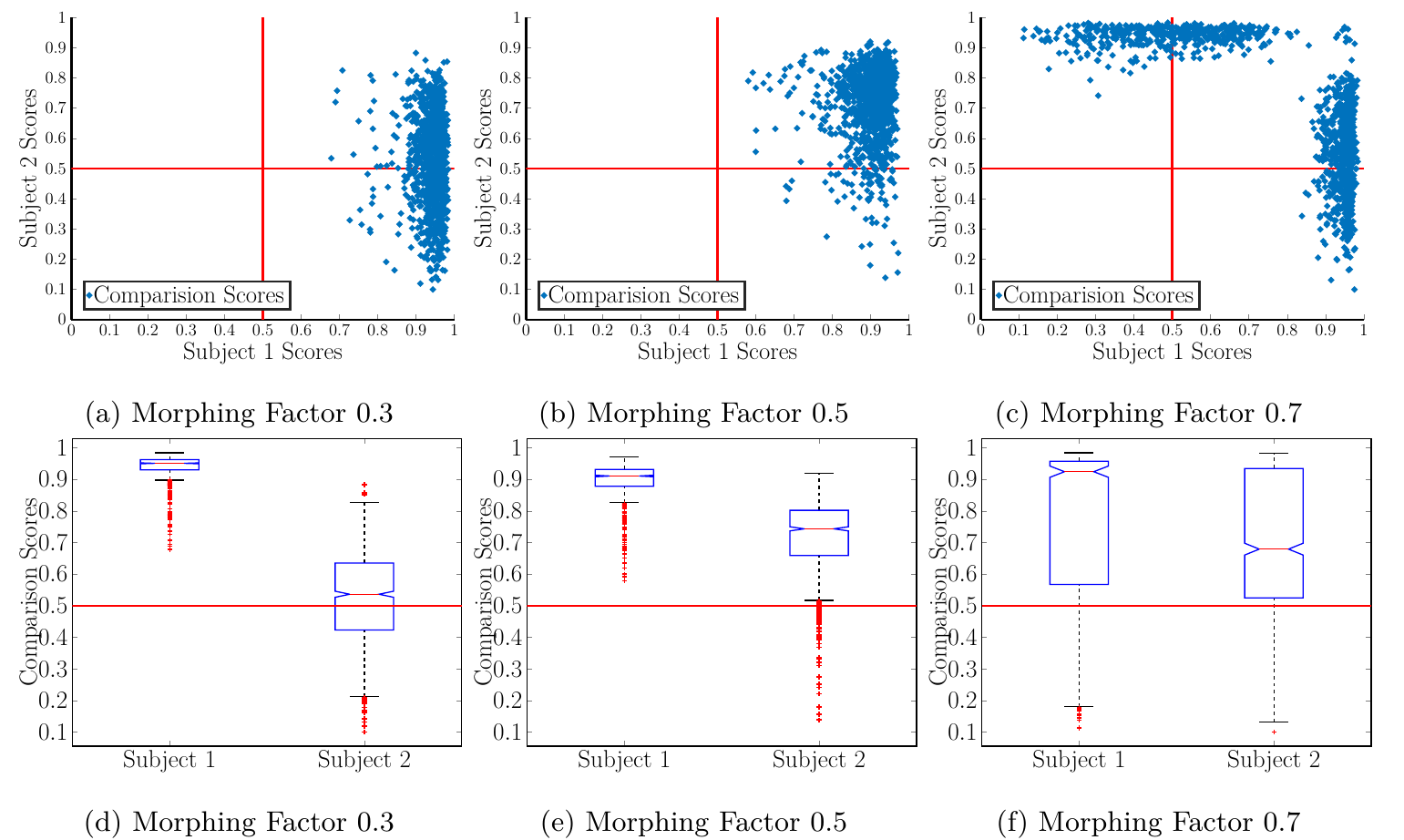}
	\vspace{-3mm}
	\caption{Scatter and Box plots obtained using COTS-I FRS on MorphAge-II dataset }
	\label{fig:CogAGII}
  \end{minipage}
  \hfill
  \begin{minipage}[b]{0.7\textwidth}
	\includegraphics[width=1\linewidth]{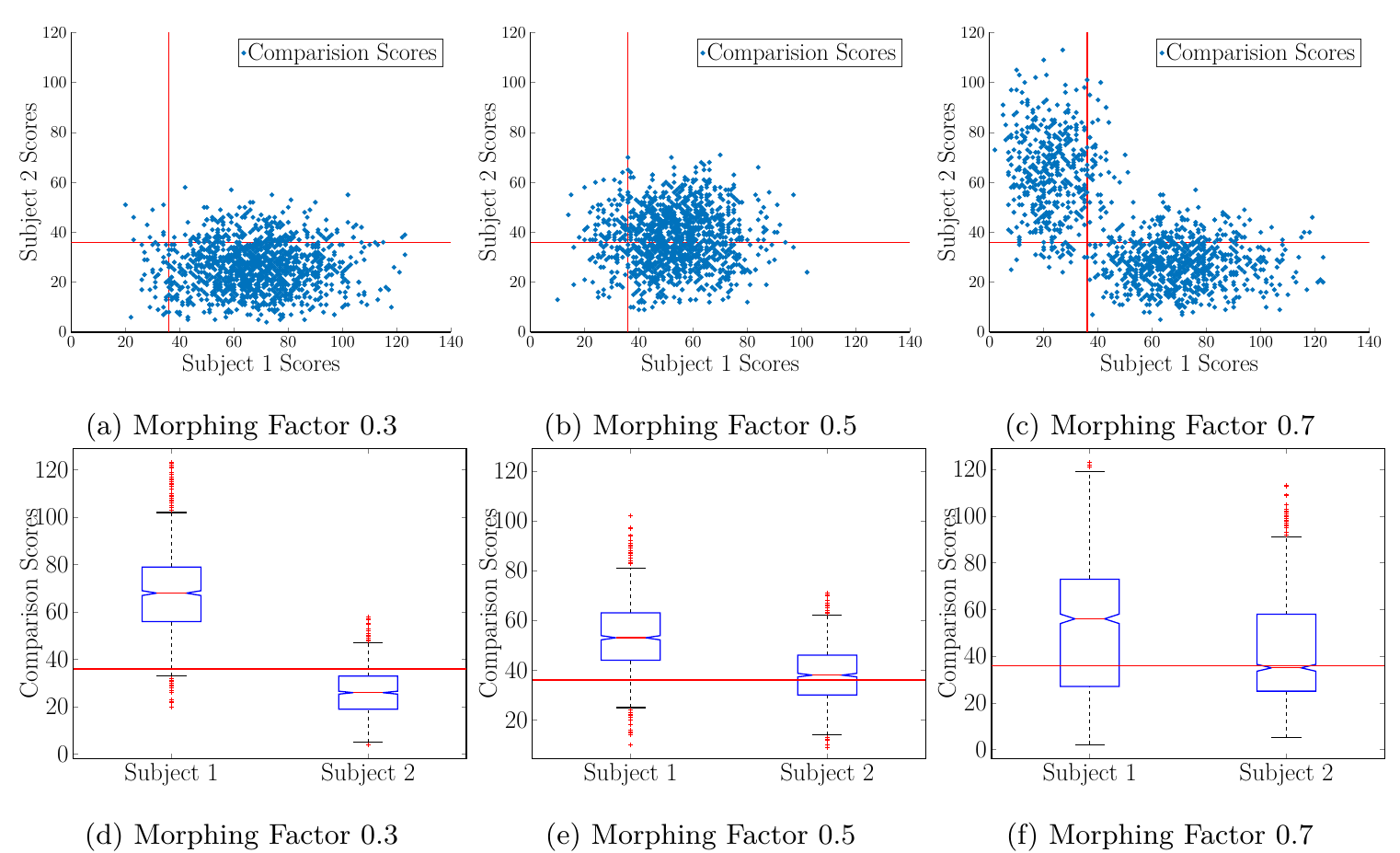}
	\vspace{-3mm}
	\caption{Scatter and Box plots obtained using COTS-II FRS on MorphAge-II dataset }
	\label{fig:NeuroAGII}
  \end{minipage}
\end{figure*}

Table \ref{tab:FMMPMR} indicates the FMMPMR ($\%$) computed using the two COTS FRS on both bins - MorphAge-I and MorphAge-II. Figure \ref{fig:CogAGI} and Figure \ref{fig:NeuAGI} shows the scatter plot and box plot for MorphAge-I dataset from two COTS FRS respectively.
Figure \ref{fig:CogAGI}(a), \ref{fig:CogAGI}(b), \ref{fig:CogAGI}(c) and  Figure \ref{fig:NeuAGI}(a), \ref{fig:NeuAGI}(b), \ref{fig:NeuAGI}(c)
provides the visualization of the comparison scores when the morphed image is enrolled, and both contributing data subjects are probed for both FRS.  In the most serve conditions, meaning a high vulnerability of the FRS with regards to morphing attacks, we will obtain comparison scores that are clustered in the top right corner of the figure.  The vertical and horizontal lines indicate the threshold that is recommended by the COTS FRS for operational settings in the border control application corresponding to $FMR=0.1\%$. 
Figure \ref{fig:CogAGI}(d), \ref{fig:CogAGI}(e), \ref{fig:CogAGI}(f) and \ref{fig:NeuAGI}(d), \ref{fig:NeuAGI}(e), \ref{fig:NeuAGI}(f)
shows the box plot that provides insight on the distributions of comparison scores corresponding to the contributor probe images allowing us to understand which of the two probe images (of the contributing subjects) are more vulnerable for the FRS. In similar lines,  Figure  \ref{fig:CogAGII} and \ref{fig:NeuroAGII} shows the scatter plot and box plot for the MorphAge-II dataset that are computed from the two COTS FRS. Based on the obtained results the following are our main observations: 
\begin{itemize}[leftmargin=*,noitemsep, topsep=0pt,parsep=0pt,partopsep=0pt]
    \item \textit{Intra-Age Groups:} As expected the morphed image with the morphing factor of 0.5 indicates the highest vulnerability as reflected by both COTS FRS.  However, the morphing factor of 0.3 and 0.7 indicates a reduced vulnerability that can be attributed to the morphing factor weights leaning toward only one of the contributing data subjects. This fact is illustrated in Figure \ref{fig:CogAGI}, \ref{fig:NeuAGI}, \ref{fig:CogAGII} and \ref{fig:NeuroAGII}, where we can observe that with a morphing factor of 0.3, the subject 1 is likely to be verified. While with a morphing factor of 0.7, in most cases, subject 2 is likely to be verified rather than subject 1. While not so surprising, the morphing factor of 0.5 indicates (almost) equally both contributing subjects can be verified. 
    \item \textit{Inter-Age groups:} Based on the obtained results, it is also interesting to note the direct influence on the morphing factor on the vulnerability. Thus, with the morphing factor of 0.3 and 0.7, both COTS FRS shows a greater reduction value of FMMPMR on MorphAge-II dataset. This indicates morphing attacks pose lesser threats to FRS under the influence of ageing. However, with the morphed factor of 0.5, the COTS-II FRS indicates lower values of FMMPMR, while COTS-I indicates a moderate reduction in the vulnerability despite being very significant. 
    \item Observing the box plots for the morphing factor of 0.5 from both MorphAge-I and MorphAge-II, it can be noted that, both the median and whiskers corresponding to the comparison scores from both subjects are reduced in MorphAge-II when compared to MorphAge-I. These observations, together with the quantitative value of FMMPMR, indicate the reduced threats to morphing attacks on FRS under ageing. This fact is consistently observed for both COTS FRS and are statistically significant as observed in the box plots. 
    \item \textit{Role of COTS FRS:} The COTS-I FRS indicates the highest vulnerability on three morphing factors when compared to that of the COTS-II FRS. The morphing factor with 0.5 shows the highest FMMPMR with $95.07\%$ on MorphAge-I and $93.81\%$ on MorphAge-II with COTS-I FRS. The lowest value of FMMPMR is noted with COTS-II FRS with a morphing factor of 0.7 in the MorphAge-II dataset. 
\end{itemize}


\begin{table}[htbp]
	\centering
	\caption{Experiment-I: Quantitative performance of the MAD techniques on MorphAge-II}
	\resizebox{1\linewidth}{!}{
		\begin{tabular}{|c|c|c|c|c|c|}
			\hline
		    \multirow{3}[6]{*}{\textbf{Algorithm}} & \textbf{Development} & \multicolumn{4}{c|}{\textbf{Testing set}} \bigstrut\\
			 &  \textbf{Set} &  \multicolumn{4}{c|}{} \bigstrut\\
            \cline{2-6} & \multirow{2}{*}{EER (\%)} & \multirow{2}{*}{EER (\%)} & \multicolumn{3}{c|}{BPCER (\%) @ APCER (\%) =} \bigstrut\\
            \cline{4-6}  &   &   & 1 & 5 & 10 \bigstrut\\
			\hline
			\multicolumn{6}{|c|}{Morphing factor ($\alpha$) 0.3} \bigstrut\\
			\hline
			LBP-SVM \cite{Raghavendra2016, Luuk_FMdetect2018, DBLP:journals/corr/abs-1901-08811, Scher2017}  & 30.64 & 29.21 & 61.24 & 48.83 & 44.96 \bigstrut\\
			\hline
			BSIF-SVM \cite{Raghavendra2016, Scher2017} & 33.35 & 39.17 & 58.91 & 51.16 & 48.83 \bigstrut\\
			\hline
			HOG-SVM \cite{Scher2017} & 32.56 & 32.56 & 66.66 & 51.93 & 45.73 \bigstrut\\
			\hline
			AlexNet-SVM \cite{DBLP:journals/corr/abs-1901-08811, RagCVIP2018, DeepLearning_JanaDittman} & 4 & 5.49 & 7.75 & 4.65 & 4.65 \bigstrut\\
			\hline
			Color Denoising \cite{Sushma_IPTA2019} & 3.15 & 1.7 & 3.1 & 0 & 0 \bigstrut\\
			\hline
			\multicolumn{6}{|c|}{Morphing factor ($\alpha$) 0.5} \bigstrut\\
			\hline
			LBP-SVM \cite{Raghavendra2016, Luuk_FMdetect2018, DBLP:journals/corr/abs-1901-08811, Scher2017}  & 28.71 & 32.39 & 68.99 & 48.06 & 41.08 \bigstrut\\
			\hline
			BSIF-SVM \cite{Raghavendra2016, Scher2017} & 32.65 & 39 & 63.56 & 51.16 & 48.83 \bigstrut\\
			\hline
			HOG-SVM \cite{Scher2017} & 30.33 & 32.56 & 62.02 & 52.71 & 44.96 \bigstrut\\
			\hline
			AlexNet-SVM \cite{DBLP:journals/corr/abs-1901-08811, RagCVIP2018, DeepLearning_JanaDittman} & 2.92 & 3.78 & 6.2 & 3.87 & 3.11 \bigstrut\\
			\hline
			Color Denoising \cite{Sushma_IPTA2019}& 3.77 & 0.75 & 1.55 & 0.77 & 0.77 \bigstrut\\
			\hline
			\multicolumn{6}{|c|}{Morphing factor ($\alpha$) 0.7} \bigstrut\\
			\hline
			LBP-SVM \cite{Raghavendra2016, Luuk_FMdetect2018, DBLP:journals/corr/abs-1901-08811, Scher2017}  & 29.33 & 27.03 & 58.91 & 51.98 & 45.73 \bigstrut\\
			\hline
			BSIF-SVM \cite{Raghavendra2016, Scher2017} & 34.7 & 33.07 & 58.91 & 51.16 & 48.06 \bigstrut\\
			\hline
			HOG-SVM \cite{Scher2017} & 31.02 & 29.09 & 73.64 & 58.91 & 44.96 \bigstrut\\
			\hline
			AlexNet-SVM \cite{DBLP:journals/corr/abs-1901-08811, RagCVIP2018, DeepLearning_JanaDittman} & 3.15 & 5.5 & 11.62 & 5.42 & 4.65 \bigstrut\\
			\hline
			Color Denoising \cite{Sushma_IPTA2019} & 3.15 & 0.75 & 3.1 & 0 & 0 \bigstrut\\
			\hline
		\end{tabular}%
	}
	\label{tab:AGII:EXP-III}%
\end{table}%
\section{Face Morph Attack Detection Performance}
\label{sec:performance}
In this section, we benchmark the most recent digital MAD techniques on the newly created MorphAge dataset. The goal of this experiment is to understand the impact of ageing on the detection performance of the MAD techniques. To this extent, we design two different experiments to reflect the variation in the performance of the MAD techniques under the influence of ageing. \textbf{Experiment-I:} the evaluation protocol is designed to evaluate the MAD detectors in the same age group. Thus, the MAD detectors are trained and tested with the same group data. \textbf{Experiment-II:} is designed to evaluate the performance of MAD detection with the variation in age. Thus, MAD detectors are trained with the MorphAge-I data and tested with only the MorphAge-II dataset. In both experiments, the corresponding development dataset is used to tune the parameters of the algorithm and also to compute the operating threshold at APCER = $1\%$, $5\%$ and $10\%$.  In this work, we have evaluated five different MAD schemes such as: Local Binary Pattern (LBP) LBP-SVM \cite{Raghavendra2016, Luuk_FMdetect2018, DBLP:journals/corr/abs-1901-08811, Scher2017}, Binarized Statistical Image Features (BSIF) \cite{Raghavendra2016, Scher2017}, Histogram of Oriented Gradients (HOG) \cite{Scher2017}, AlexNet \cite{DBLP:journals/corr/abs-1901-08811, RagCVIP2018, DeepLearning_JanaDittman} and Color Denoising \cite{Sushma_IPTA2019}. We have considered these five MAD techniques as they have indicated good performance on three different large scale digital morphing datasets \cite{Sushma_IPTA2019}.  The quantitative results are presented according to the ISO/IEC 30107-3 \cite{ISO/IEC2015a} metrics such as Bona fide Presentation Classification Error Rate (BPCER($\%$)) and Attack Presentation Classification Error Rate (APCER ($\%$)) along with D-EER(\%).

Table \ref{tab:AGI_EXP-III} and Table \ref{tab:AGII:EXP-III} indicates the quantitative results of the MAD schemes on two different age groups MorphAge-I and MorphAge-II respectively on the Experiment-I protocol. Based on the obtained results, it can be noticed that: 
\begin{itemize}[leftmargin=*,noitemsep, topsep=0pt,parsep=0pt,partopsep=0pt]
    \item The traditional MAD methods based on LBP, BSIF, and HOG fail to indicate acceptable detection performance for both MorphAge-I and MorphAge-II dataset.
    \item Recently introduced MAD techniques based on AlexNet and Color denoising techniques have shown excellent performance in detecting morphing attacks. 
    \item It is interesting to note that the MAD methods do not show any influence of the different morphing factors on the detection performance. The detection performance with different morphing factor did further not vary irrespective of the age group as well.
    \item Among the five benchmarked different MAD techniques, the color denoising MAD has indicated the best performance across various morphing factors ($\alpha$) for both MorphAge-I and MorphAge-II. 
\end{itemize}


\begin{table}[htbp]
  \centering
  \caption{Experiment-II: Quantitative detection performance of MAD techniques on MorphAge-I v/s. MorphAge-II}
   \resizebox{0.95\linewidth}{!}{
    \begin{tabular}{|c|c|c|c|c|c|}
    \hline
    \multirow{3}[6]{*}{\textbf{Algorithm}} & \textbf{Development} & \multicolumn{4}{c|}{\textbf{Testing set}} \bigstrut\\
	&  \textbf{Set} &  \multicolumn{4}{c|}{} \bigstrut\\
    \cline{2-6} & \multirow{2}{*}{EER (\%)} & \multirow{2}{*}{EER (\%)} & \multicolumn{3}{c|}{BPCER (\%) @ APCER (\%) =} \bigstrut\\
    \cline{4-6}  &   &   & 1 & 5 & 10 \bigstrut\\
    \hline
    \multicolumn{6}{|c|}{Morphing factor ($\alpha$) 0.3} \bigstrut\\
    \hline
    LBP-SVM \cite{Raghavendra2016, Luuk_FMdetect2018, DBLP:journals/corr/abs-1901-08811, Scher2017}  & 28.14 & 34.19 & 92.24 & 65.89 & 47.28 \bigstrut\\
    \hline
    BSIF-SVM \cite{Raghavendra2016, Scher2017} & 31.82 & 44.13 & 100 & 98.44 & 84.49 \bigstrut\\
    \hline
    HOG-SVM \cite{Scher2017} & 32.09 & 41.86 & 91.47 & 70.54 & 62.01 \bigstrut\\
    \hline
    AlexNet-SVM \cite{DBLP:journals/corr/abs-1901-08811, RagCVIP2018, DeepLearning_JanaDittman} & 4.38 & 3.03 & 8.52 & 3.10 & 2.32 \bigstrut\\
    \hline
    Color Denoising \cite{Sushma_IPTA2019} & 1.63 & 2.27 & 1.55 & 0.45 & 0 \bigstrut\\
    \hline
    \multicolumn{6}{|c|}{Morphing factor ($\alpha$) 0.5} \bigstrut\\
    \hline
    LBP-SVM \cite{Raghavendra2016, Luuk_FMdetect2018, DBLP:journals/corr/abs-1901-08811, Scher2017}  & 27.82 & 33.29 & 86.04 & 66.66 & 48.06 \bigstrut\\
    \hline
    BSIF-SVM \cite{Raghavendra2016, Scher2017} & 31.82 & 45.42 & 100 & 96.89 & 84.49 \bigstrut\\
    \hline
    HOG-SVM \cite{Scher2017} & 30.73 & 37.95 & 85.27 & 67.44 & 57.36 \bigstrut\\
    \hline
       AlexNet-SVM \cite{DBLP:journals/corr/abs-1901-08811, RagCVIP2018, DeepLearning_JanaDittman} & 3.18 & 0.94 & 3.10 & 0.77 & 0.39 \bigstrut\\
    \hline
    Color Denoising \cite{Sushma_IPTA2019}& 1.63 & 1.59 & 0.7 & 0 & 0 \bigstrut\\
    \hline
    \multicolumn{6}{|c|}{Morphing factor ($\alpha$) 0.7} \bigstrut\\
    \hline
    LBP-SVM \cite{Raghavendra2016, Luuk_FMdetect2018, DBLP:journals/corr/abs-1901-08811, Scher2017}  & 28.86 & 32.24 & 93.20 & 65.89 & 49.61 \bigstrut\\
    \hline
    BSIF-SVM \cite{Raghavendra2016, Scher2017} & 31.90 & 37.33 & 100 & 96.89 & 84.49 \bigstrut\\
    \hline
    HOG-SVM \cite{Scher2017} & 32.98 & 33.54 & 88.37 & 68.99 & 58.13 \bigstrut\\
    \hline
       AlexNet-SVM \cite{DBLP:journals/corr/abs-1901-08811, RagCVIP2018, DeepLearning_JanaDittman} & 5.08 & 2.27 & 6.97 & 3.87 & 0.77 \bigstrut\\
    \hline
    Color Denoising \cite{Sushma_IPTA2019} & 2.75 & 2.46 & 3.10 & 0.40 & 0 \bigstrut\\
    \hline
    \end{tabular}%
    }
  \label{tab:AGI:AGII}%
\end{table}%

Table \ref{tab:AGI:AGII} indicates the quantitative detection performance of MAD methods in Experiment-II.  Based on the obtained results, it can be noted that the ageing does not influence the performance of the MAD methods. It is worth noting that, in this protocol, MAD methods are trained using only MorphAge-I dataset and are tested on the MorphAge-II dataset with the age difference up to 5 years. Further, the data subjects in MorphAge-I and MorphAge-II do not overlap. Among the five different MAD methods, color denoising based MAD has again indicated the best performance for all three morphing factors ($\alpha$). 
As it can be deduced, ageing does not influence the detection capabilities of MAD under the performed experimental settings. 

\section{Discussion}
Based on the observations made above from the experiments and obtained results, the research questions formulated in Section \ref{sec:FAMA} are answered below.
\begin{itemize}[leftmargin=*,noitemsep, topsep=0pt,parsep=0pt,partopsep=0pt]
    \item {Q1. How vulnerable are COTS FRS when a composite morph image is enrolled and is after a period of ageing probed against a live image from one of the contributing subjects?} 
    \begin{itemize}[leftmargin=*,noitemsep, topsep=0pt,parsep=0pt,partopsep=0pt]
        \item Supported by the obtained experimental results reported in Table \ref{tab:FMMPMR}, it is interesting to note that the value of FMMPMR is reduced to certain extent in case of MorphAge-II dataset. The morphed images are not easily verified against the probe images after a certain degree of ageing making FRS less vulnerable. 
    \end{itemize}
    \item {Q2. Do current Morphing Attack Detection (MAD) algorithms scale-up to detect such attacks under the influence of ageing?} 
        \begin{itemize}[leftmargin=*,noitemsep, topsep=0pt,parsep=0pt,partopsep=0pt]
            \item Based on the experimental results reported in Table \ref{tab:AGII:EXP-III}, ageing has negligible impact on the MAD and thereby the existing MAD schemes can detect the attacks even under ageing.
        \end{itemize}
    \item {Q3. What is the impact of different alpha (or blending, morphing) factors used to generate the morphed image under the constraint of ageing, specifically with respect to MAD?} 
        \begin{itemize}[leftmargin=*,noitemsep, topsep=0pt,parsep=0pt,partopsep=0pt]
            \item Based on the experimental results, it is interesting to note that the morphing factors $alpha=0.3$ and  $0.7$ show greater reduction in the vulnerability in both the COTS FRS with respect to ageing as reported in Table \ref{tab:FMMPMR}. It has to be however noted that COTS-II FRS indicates lower vulnerability when a morphing factor of 0.5 is employed.
        \end{itemize}
\end{itemize}


    


\section{Conclusion}
\label{sec:conclusion}

We have presented an empirical study on quantifying the vulnerability of COTS FRS with regards to morphing attacks under the influence of ageing. We have introduced a new dataset with two different age groups derived from the publicly available MORPH II face dataset referred as MorphAge-I and MorphAge-II. Further, we have also introduced a new evaluation metric namely, Fully Mated Morphed Presentation Match Rate(FMMPMR) to quantify the vulnerability effectively. Extensive experiments were carried out using two different COTS FRS and three different morphing factors(with $\alpha$ = 0.3, 0.5 and 0.7). Based on the obtained results, it is observed that impact of ageing reduces the vulnerability from morphing attacks on COTS FRS. The reduction in the vulnerability is more prominent when the morphing factor is $\alpha$ = 0.3 and 0.7. However with a morphing factor of $\alpha$ = 0.5, the vulnerability does not change significantly with the COTS-I, while COTS-II FRS still indicates a significant reduction in the vulnerability. Extensive experiments were performed to quantify the performance variation of the MAD methods under the influence of ageing. To this extent, three different evaluation protocols are presented that show no influence of ageing on morph attack detection performance. It is also interesting to note that robust MAD methods are not sensitive to variations of the morphing factor even under the influence of ageing. 



%
\balance
{\footnotesize
\bibliographystyle{ieee}
\bibliography{morphing-ageing-impact-191214}
}


\end{document}